 \documentclass[10pt,twocolumn,letterpaper]{article} 
 \usepackage{cvpr} 
\usepackage{amsmath}
\usepackage{colortbl}           
\usepackage{algorithm,algorithmic}
\usepackage{multirow}
\usepackage{xcolor}
\usepackage{graphics}
\usepackage{amssymb}
\usepackage{natbib}
\setcitestyle{square,sort,comma,numbers}
\usepackage{epsfig}
\usepackage{flushend}
\usepackage{tabularx}
\usepackage{epstopdf}
\usepackage[switch]{lineno}
\usepackage{caption}
\usepackage{changepage}
\usepackage{color}
\usepackage{url}
\usepackage{booktabs}
\usepackage{pifont}
\definecolor{mygray}{gray}{.9}

\newcommand{\yes}{\color{blue}{\ding{51}}}
\newcommand{\no}{\color{red}{\ding{55}}}

\definecolor{cvprblue}{rgb}{0.21,0.49,0.74}
\usepackage[pagebackref,breaklinks,colorlinks,allcolors=cvprblue]{hyperref}

\title{CoT4AD: A Vision-Language-Action Model with Explicit Chain-of-Thought Reasoning for Autonomous Driving}

\author{Zhaohui Wang \quad Tengbo Yu \quad Hao Tang$^*$ \\
Peking University\\
$^*${corresponding author: bjdxtanghao@gmail.com}
}

\begin{document}
\maketitle

\begin{abstract}

 Vision-Language-Action (VLA) models have recently attracted growing attention in end-to-end autonomous driving for their strong reasoning capabilities and rich world knowledge. However, existing VLAs often suffer from limited numerical reasoning ability and overly simplified input–output mappings, which hinder their performance in complex driving scenarios requiring step-by-step causal reasoning. To address these challenges, we propose \textbf{CoT4AD}, a novel VLA framework that introduces \textbf{C}hain-\textbf{o}f-\textbf{T}hought (CoT) reasoning for \textbf{a}utonomous \textbf{d}riving to enhance both numerical and causal reasoning in Vision-Language Models (VLMs). CoT4AD integrates visual observations and language instructions to perform semantic reasoning, scene understanding, and trajectory planning. During training, it explicitly models a perception–question–prediction–action CoT to align the reasoning space with the action space across multiple driving tasks. During inference, it performs implicit CoT reasoning to enable consistent numerical reasoning and robust decision-making in dynamic environments. Extensive experiments on both real-world and simulated benchmarks, including nuScenes and Bench2Drive, demonstrate that CoT4AD achieves state-of-the-art performance in both open-loop and closed-loop evaluations. Code will be released upon paper acceptance.

\end{abstract}

\begin{figure*}[!t]
	\centering
	\includegraphics[width=1\linewidth]{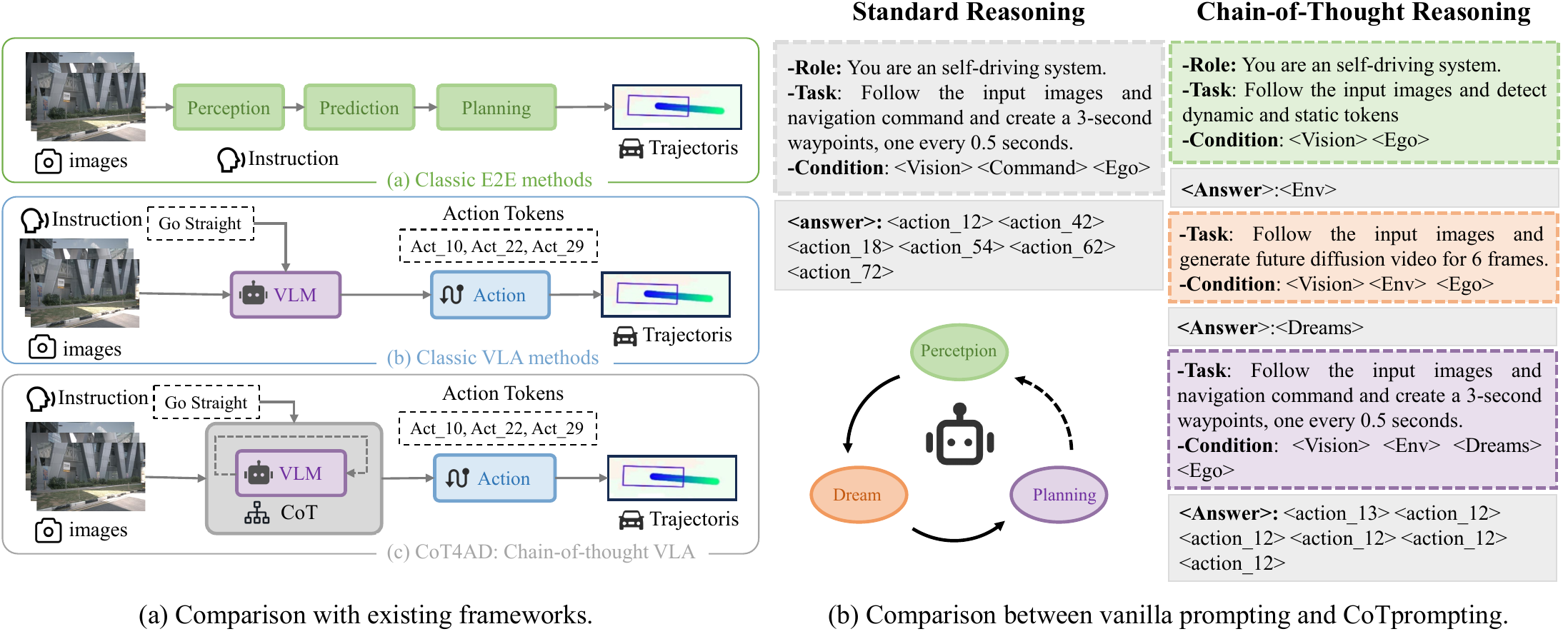}
	\caption{{{
				\textbf{Comparison with existing frameworks.} (a) Classic E2E methods map sensor inputs to control outputs with modular designs. (b) Classic VLA methods incorporate language reasoning and VLMs to map sensor inputs to action space for planning. (c) CoT4AD (Outs) incorporate CoT reasoning for VLMs to enable explicit mult-step reasoning of planning.
	}}}
	\label{fig:compare}
	\vspace{-0.5cm}
\end{figure*}

\section{Introduction}

Autonomous driving, as a core research direction in artificial intelligence and robotics, has attracted widespread attention in recent years. It not only holds the potential to improve traffic safety and travel efficiency but also plays a key role in the development of smart cities and intelligent transportation systems \cite{wang2010parallel,jin2023intelligent}. Traditional autonomous driving systems typically adopt a modular pipeline architecture, decomposing perception \cite{qian2022badet,li2024bevformer}, prediction \cite{chambon2024pointbev,sirko2024occfeat} and planning \cite{ma2015efficient} into separate modules. However, such methods often suffer from error accumulation, difficulties in cross-module optimization, and limited generalization ability in practical applications, which constrain the performance of autonomous driving systems in complex environments \cite{hu2023planning}.

To address these challenges, end-to-end autonomous driving paradigms have been proposed. These methods aim to leverage a unified learning framework to directly predict driving control signals or plan trajectories from raw sensor inputs, thereby avoiding the uncertainties introduced by multi-stage information propagation \cite{chen2024end}. Meanwhile, with the rapid development of large-scale Vision-Language Models (VLMs), researchers have begun to explore the potential of Vision-Language-Action (VLA) models \cite{kim2024openvla,zitkovich2023rt} for end-to-end autonomous driving. VLA models are capable of handling multi-modal inputs and performing semantic reasoning through language instructions, demonstrating stronger interpretability and generalization capabilities compared with traditional end-to-end approaches \cite{liu2024survey,zhou2024gpt}.

Through imitation learning on diverse datasets, VLAs inherit the strong ability of VLMs to understand heterogeneous scenes, objects, and language, enabling robust cross-task generalization. However, they also share VLMs’ weaknesses—particularly poor numerical reasoning in complex environments \cite{zhou2025exploring,frieder2023mathematical}. Thus, applying VLMs to autonomous driving faces two key challenges: (1) limited numerical reasoning causes unreliable or hallucinatory predictions and (2) existing methods treat LLMs as monolithic mappers from perception to numerical outputs, neglecting their capability for multi-step reasoning.

Chain-of-Thought (CoT) reasoning enhances LLM reasoning by decomposing complex tasks into intermediate steps \cite{wei2022chain,guo2025ifship}. While prior works introduce CoT via language descriptions, keypoints, or bounding boxes \cite{du2023learning,mu2023embodiedgpt}, they mainly focus on embodied or robotic settings with constrained environments \cite{zhao2025cot,ni2024generate}. However, autonomous driving presents fundamentally different challenges—it demands accurate numerical reasoning, long-horizon planning, and robust generalization in dynamic, large-scale and safety-critical environments. As shown in Figure~\ref{fig:compare}(b), directly feeding all prompts into a VLM leads to unstable outputs, whereas incorporating CoT enables structured intermediate reasoning and more reliable results. This work try to address a key question: \textbf{How can CoT reasoning be tailored for autonomous driving to enhance efficiency and performance?}

To this end, we propose CoT4AD, a novel VLA model that integrates CoT reasoning for end-to-end autonomous driving. As illustrated in Figure \ref{fig:compare}, CoT4AD differs from previous methods by finetuning open-source pretrained language models through a series of downstream tasks tailored for autonomous driving scenarios, while being able to explicitly or implicitly perform CoT reasoning. This enables the model to generate reliable and stable trajectories from multi-modal inputs. 

Our unified framework integrates environmental perception, language reasoning, future prediction, and trajectory planning, enabling the model to produce explicit CoT reasoning steps.
In perception reasoning, the model is trained on perception tasks using expert data to acquire scene understanding capabilities.
In future prediction, the model learns to predict future scenes from expert data, enabling the reconstruction of these scenes.
In VQA reasoning, the model is fine-tuned on Visual Question Answering (VQA) tasks using prompt-based supervision.
In trajectory planning, the model performs imitation learning to generate high-quality driving trajectories.
Through a multi-stage training process, CoT4AD develops CoT reasoning capabilities tailored to autonomous driving scenarios. During inference, CoT4AD can directly produce driving trajectories in a single forward pass without explicitly generating intermediate reasoning steps, achieving a balance between planning performance and computational efficiency.

We extensively evaluate CoT4AD using real-world dataset nuScenes \cite{caesar2020nuscenes} and simulation dataset Bench2Drive \cite{jia2024bench2drive}. Experimental results demonstrate that CoT4AD achieves superior performance across various end-to-end autonomous driving benchmarks under both open-loop and closed-loop tests. 

The main contributions are summarized as follows:

\begin{enumerate}[(1)]
    \item We introduce CoT4AD, an end-to-end autonomous driving framework leveraging a pretrained VLM with multi-step finetunes, enabling CoT reasoning and mult-task capabilities from raw visual observations and language instructions.

    \item We introduce an innovative approach to future prediction and trajectory planning for driving scenes. This diffusion-based framework integrates existing video generation and planning approaches and is seamlessly integrated CoT reasoning pipelines.

    \item Extensive experiments on NuScenes and Bench2Drive datasets show that CoT4AD establishes new state-of-the-art results in both open-loop and closed-loop driving, consistently outperforming prior LLM-based and end-to-end autonomous driving approaches.
\end{enumerate}

\section{Related Work}\label{sec:relate}
\subsection{End-to-End Autonomous Driving (E2E-AD)}
E2E-AD directly predicts driving trajectories from raw sensor data without decomposing the pipeline into perception, prediction, and planning. Early works such as ALVINN \cite{pomerleau1988alvinn} and PilotNet \cite{bojarski2016end} pioneered this idea but suffered from poor generalization and interpretability in complex scenes.
With large-scale datasets \cite{caesar2020nuscenes,sun2020scalability} and advanced architectures \cite{vaswani2017attention}, methods such as CIL \cite{codevilla2018end}, TransFuser \cite{chitta2022transfuser}, and LAV \cite{chen2022learning} improve behavioral control and robustness through multimodal fusion and spatial–temporal reasoning.
Recent trajectory-based approaches \cite{hu2023planning,jiang2023vad} jointly optimize perception, prediction, and planning for better safety and interpretability. Meanwhile, VLM-driven systems such as DriveGPT \cite{huang2024drivegpt} and Talk2Drive \cite{cui2023large} enable reasoning and explainability via natural language, and diffusion-based models \cite{liao2025diffusiondrive,jiang2025diffvla} capture trajectory uncertainty.
Despite these advances, E2E-AD still struggles with robustness and interpretability in real-world scenarios.

\subsection{Vision Language Models (VLMs)}
VLMs aim to jointly model images and natural language, achieving a semantic understanding of images and text through cross-modal alignment learning. Early works such as VisualBERT \cite{li2019visualbert} and ViLT \cite{kim2021vilt} extended the BERT architecture to vision-language tasks, enabling unified modeling for image-text matching and visual question answering (VQA) through large-scale joint pretraining on paired image-text datasets. With the rise of Multimodal Large Language Models (MLLMs), VLMs have gradually acquired strong reasoning, comprehension, and generative capabilities. Models like Flamingo \cite{alayrac2022flamingo} and LLaVA \cite{liu2023visual} further integrate large language models (LLMs) with visual encoders, allowing for contextual understanding of complex scenes and supporting open visual question answering, caption generation, and reasoning tasks. Recently, general-purpose VLMs such as GPT-4V \cite{achiam2023gpt}and Qwen-VL \cite{bai2023versatile} have demonstrated remarkable cross-task generalization, effectively solving diverse vision-language understanding and reasoning tasks without task-specific fine-tuning.

\subsection{Chain-of-Thought (CoT) Reasoning}
CoT reasoning has recently emerged as a powerful paradigm for enhancing the reasoning capabilities of LLMs. Its core idea is to make the model explicitly generate intermediate reasoning steps before producing the final answer, thereby decomposing complex problems into a series of interpretable intermediate processes \cite{wei2022chain}. CoT reasoning has been proven effective in improving models’ numerical and logical reasoning abilities. 
In robotic domains, introducing CoT reasoning offers a new perspective for decision-making. CoT-VLA \cite{zhao2025cot} predicts subgoal images as intermediate reasoning step and then generates action tokens to accomplish manipulation tasks. ECoT \cite{zawalski2024robotic} increases the absolute success rate of OpenVLA \cite{kim2024openvla} by performing multiple reasoning steps before task execution. Although CoT reasoning has shown great potential in robotic applications, its integration into VLAs for autonomous driving remains in the early stages of exploration.

\begin{figure*}[!htb]
	\centering
	\includegraphics[width=\textwidth]{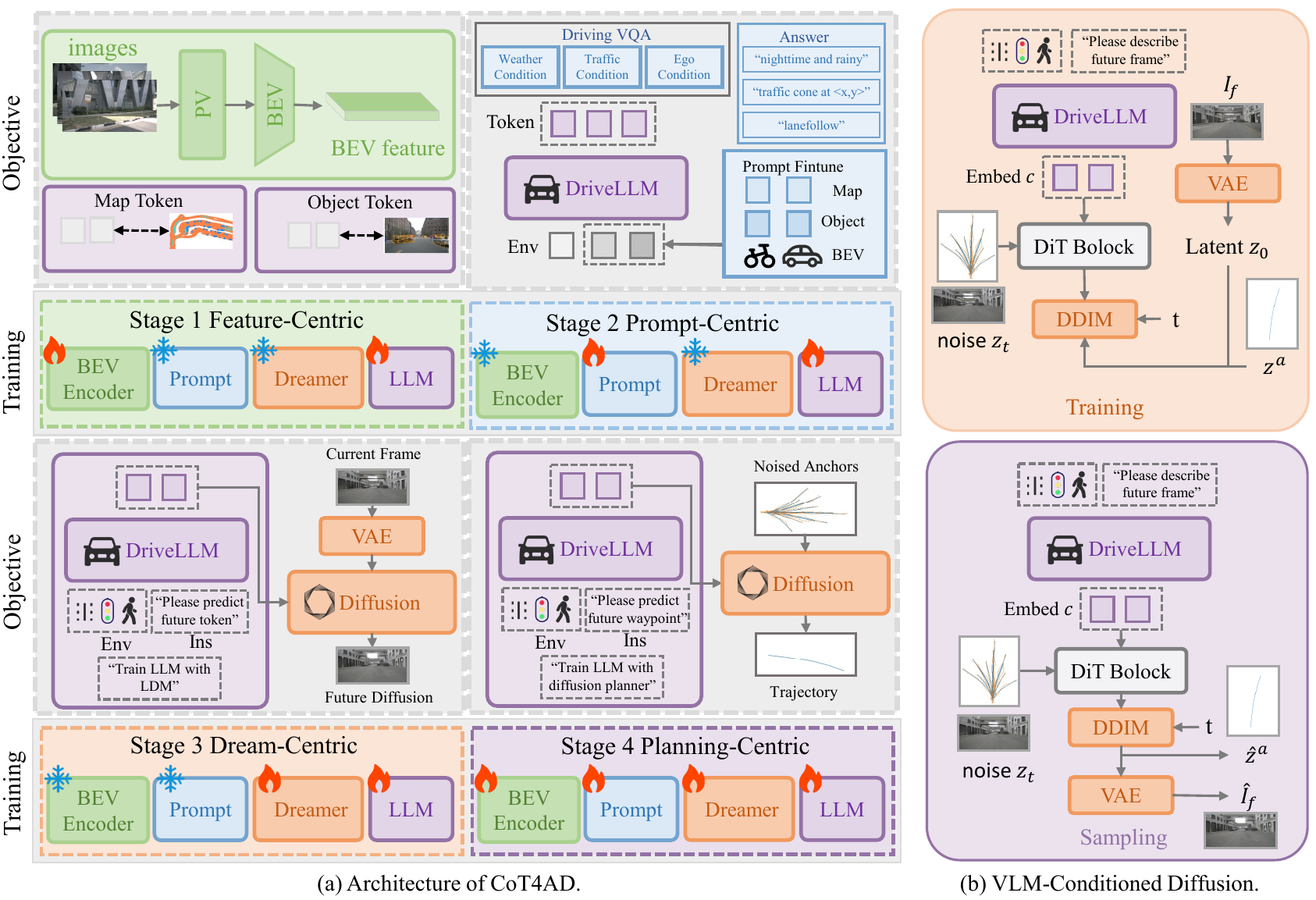}
	\caption{{(a) Architecture of CoT4AD. It consists of four stages of CoT reasoning including: 3D perception, VQA, VLM-conditioned diffusion and planning. (b) VLM-Conditioned Latent Diffusion. A conditional latent DiT model diffuses the latent of the current frame conditioned on VLM embeddings, and the future frame is reconstructed via a VAE decoder.}}
	\label{fig:arch}
\end{figure*}

\section{The Proposed Method}\label{sec:method}

\subsection{3D Environmental Perception}\label{BEV Encoder}

In existing VLAs, visual features are typically extracted using 2D encoders \cite{kim2024openvla,zitkovich2023rt}, which perform well in planar tasks \cite{o2024open} but struggle to model 3D structures and spatial relationships. Due to the lack of multi-view depth and geometric consistency, these models are unbale to accurately perceive 3D environments. To address this limitation, CoT4AD employs feature-centric perception training, learning from multiple 3D perception tasks to generate 3D visual tokens.

As shown in Figure~\ref{fig:arch} (a), CoT4AD takes multi-view images $I=\{I^i\}_{i=1}^N$ as input. A 2D backbone extracts multi-scale features $f_{2D}$, which are projected into BEV space using camera parameters to obtain BEV features $f_{BEV}$. The PB-SSM framework is used for efficient multi-view fusion. Two visual modules then encode structured 3D semantics: MapTokenizer $T_{map}$ for static elements (lanes, drivable areas) and ObjectTokenizer $T_{obj}$ for dynamic objects (vehicles, pedestrians). $T_{map}$ produces map tokens $v_{map}$ from BEV patches, while $T_{obj}$ applies ROI Align \cite{he2017mask} on $f_{BEV}$ to yield object tokens $v_{obj}$. To complement limited annotation-based learning, we further introduce BEVTokenizer, which directly partitions BEV features into $N_{bev}\times N_{bev}$ patches to generate comprehensive BEV tokens $v_{bev}$. The final environmental representation is $\mathbf{V}_{env}=\{v_{map},v_{obj},v_{bev}\}$, serving as the perception stage for CoT reasoning.

\subsection{Vision-Language Prompt Tuning}\label{Env Former}
In existing VLAs, vision-language tuning is often limited to the adjustment of image tokens and language tokens \cite{kim2024openvla}, lacking joint optimization of perception tokens and other multi-modal representations. To address the limited environment representation capacity of multi-feature tokens extracted by perception models and the discretized embeddings of natural language, we introduce a VQA-based multi-modal vision-language finetuning framework.

During the fine-tuning stage, the model learns high-level perception capabilities and driving knowledge from VQA datasets, enabling the transfer from multi-modal tokens to numerical reasoning space. Inspired by soft prompt tuning~\cite{philippy2024soft}, we introduce stage-irrelevant tokens denoted as $\mathbf{V}_{s}$. These learnable and discretized tokens are used to encode visual details during training and serve as input across different stages of CoT reasoning. The effectiveness of soft prompt tuning has been demonstrated in various Transformer-based tasks, including image classification and numerical reasoning \cite{DarcetOMB24,bu2025univla}. Our experiments will show that this method is also effective for end-to-end autonomous driving tasks to improve the integration of multi-modal perception and language reasoning.

As illustrated in Figure~\ref{fig:arch} (b), we fomulate the VQA datasets as prompt-response pairs $\{\mathbf{X}_{input},\mathbf{X}_{answer}\}$, where $\mathbf{X}_{input} = \{\mathbf{V_{env}},\mathbf{V_s},\mathbf{V_{ego}}\}$, $\mathbf{V}_{ego}$ encodes the ego vehicle state such as velociy, acceleration and yaw. During instruction tuning, visual tokenizers are frozen while the LLM is set trainable. The Vision-language prompt tuning state is as:
\begin{equation}
    \hat{\mathbf{X}}_{answer}=\mathrm{LLM}\left(\mathbf{V}_{env},\mathbf{V}_{s},\mathbf{V}_{ego}\right).
\end{equation}

\subsection{VLM-Conditioned Latent Diffusion }\label{World Dreamer}

Existing VLA systems are often constrained by text-level reasoning, overlooking the rich multimodal characteristics of the real world. Inspired by world models in autonomous driving, which learn physical laws by generating future frames, we leverage a VLM-conditioned diffusion model to generate high-fidelity future frames, enabling a deeper understanding of the world’s rich semantic information.
As illustrated in the Figure \ref{fig:arch} (b), our proposed VLM-conditioned diffusion model operates as follows:

Following Latent diffusion models(LDMs) \cite{rombach2022high}, we train an encoder $E$ to compresses images $I_f$ into smaller spatial representations $z_0=E(I_f)$ and an decoder $D$ to restore future frame $I_f=D(z_0)$.  By performing diffusion modeling in the latent space, we avoid direct diffusion in the high-dimensional pixel space, thereby significantly accelerating the training process of the diffusion model. The LDM poses a forward process as gradually adding noise to the latent space, which can be defined as:
\begin{equation}
q\left(z_t\mid z_0\right)=\mathcal{N}\left(z_t;\sqrt{\bar{\alpha_t}}\boldsymbol{z_0},\left(1-\bar{\alpha_t}\right)\mathbf{I}\right),
\end{equation}
where $z_0$ is the clean latent variable, and $z_t$ is the noise frame at time t. The constants $\alpha_t$ are hyperparameters. For the reverse process, we train a diffusion transformer $f_\theta$ to reconstruct $z_0$ from the noise frame $z_t$ and the conditional embedding $c$, where $c=\mathrm{LLM}\left(\mathbf{V}_{env},\mathbf{V}_{s},\mathbf{V}_{ego}\right)$. Instead of sampling noise from a Gaussian distribution, we formulate the noise frame $z_t$ from the current frame $I_c$:
\begin{equation}
z_i=\sqrt{\bar{\alpha_i}}D(I_c)+\sqrt{1-\bar{\alpha}_i}\boldsymbol{\epsilon},\quad\boldsymbol{\epsilon}\thicksim\mathcal{N}(0,\mathbf{I}),
\end{equation}
where $i\in[1,T]$.
During training, the diffusion transformer $f_\theta$ takes noise frame $z_t$ and conditional embedding $c$ as input to predict denoised latent variable $\hat z_0=f_\theta(z_0,c)$. The training objective combines latent variable reconstruction and noise prediction losses:  
\begin{equation}
\mathcal{L}=\lambda\|z_0-\hat{z}_0\|_2^2+\lambda_\epsilon\|\epsilon-\hat{\epsilon}_\theta(z_t,c)\|_2^2.
\end{equation}

During inference, we adopt a truncated denoising process  starting from a noisy latent representation $z_t$ sampled from the current frame-based Gaussian distribution and progressively denoise it conditioned on the conditional embedding $c$ from VLM to obtain the final predicted image $\hat z_0$.
At each denoising step, the DDIM update rule is applied to iteratively refine the latent state toward the next time step. Decoder $D$ is applied to decode $z_0$ to predicted future frame $\hat I_f=D(z_0)$.

Through this latent space VLM-conditioned diffusion generation approach, VLA model can learn future scene prediction, thereby developing a form of visual reasoning about scene changes and enhancing the understanding of environmental semantics and physical laws, rather than relying solely on text or single-frame information.

\begin{table*}[htbp!] \small
\centering
\caption{
Results of E2E-AD methods on NuScenes. \dag: The ego status and planning trajectory are both processed by LLM in textual modality. ${}\ddagger$: The high-level command is not used during the training and testing phases.}
\label{tab: nuscene results}
\resizebox{0.85\textwidth}{!}
{
    \begin{tabular}{l cc cc  cccc  cccc}
    \toprule
    \multirow{2.2}{*}{Method} &
    \multirow{2.2}{*}{Reference} &
    \multirow{2.2}{*}{VLM-Based} &
    \multicolumn{2}{c}{Ego Status} &
    \multicolumn{4}{c}{L2 (m) $\downarrow$} &
    \multicolumn{4}{c}{Collision (\%) $\downarrow$} \\
    \cmidrule(lr){4-5} 
    \cmidrule(lr){6-9} 
    \cmidrule(lr){10-13}
    & & & BEV & Planner 
    & 1s & 2s & 3s & Avg. 
    & 1s & 2s & 3s & Avg. \\
    \midrule
    ST-P3~\cite{hu2022st} & ECCV’22 &- &- &- & 1.33 & 2.11 & 2.90 & 2.11 & 0.23 & 0.62 & 1.27 & 0.71 \\
    UniAD~\cite{hu2023planning} & CVPR'23&- &-  & - & 0.48 & 0.96 & 1.65 & 1.03 & 0.05 & 0.17 & 0.71 & 0.31 \\
    UniAD~\cite{hu2023planning} & CVPR'23 & -&\checkmark& \checkmark & 0.20 & 0.42 & 0.75& 0.46 & 0.02 & 0.25 & 0.84&0.37  \\
    
    VAD-Base~\cite{jiang2023vad}& CVPR'23 &- &-&-& 0.69 & 1.22 & 1.83 &1.25 & 0.06 & 0.68 & 2.52 &1.09  \\
    VAD-Base~\cite{jiang2023vad}& CVPR'23 &-&\checkmark&-& 0.41 & 0.70 & 1.06&0.72 & 0.04 & 0.43 & 1.15 &0.54\\
    VAD-Base~\cite{jiang2023vad} & CVPR'23 &-&\checkmark& \checkmark& 0.17 & 0.34 & 0.60 &0.37 & 0.04 & 0.27 & 0.67 & 0.33  \\
    \midrule
    Ego-MLP~\cite{zhai2023rethinking} &CVPR'24&- & - & \checkmark& 0.15 & 0.32 & 0.59  & 0.35&0.00 & 0.27 & 0.85&0.37\\
    BEV-Planner~\cite{li2024ego}&CVPR'24&- &- &- & 0.30 & 0.52&0.83 &0.55 & 0.10 & 0.37 & 1.30 &0.59\\
    BEV-Planner++~\cite{li2024ego} &CVPR'24&- &\checkmark &\checkmark & 0.16 & 0.32& 0.57 & 0.35& 0.00 & 0.29 & 0.73 &0.34 \\
    
    \midrule
    DriveVLM\textdagger~\cite{tian2024drivevlm}&ECCV'24& \checkmark &- & - &0.18 &0.34 &0.68 &0.40 &0.10 &0.22 &0.45 &0.27 \\
    DriveVLM-Dual~\cite{tian2024drivevlm}&ECCV'24& \checkmark &\checkmark &- & 0.15 & 0.29 & 0.48 & 0.31 & 0.05 & 0.08 & \textbf{0.17} & \textbf{0.10} \\
    
    OmniDrive$\ddagger$~\cite{wang2024omnidrive}&CoRR'24& \checkmark &-&-& 1.15 & 1.96 & 2.84 & 1.98 & 0.80 & 3.12 & 7.46 & 3.79 \\
    OmniDrive~\cite{wang2024omnidrive} &CoRR'24& \checkmark &-&-& 0.40 & 0.80 & 1.32 & 0.84 & 0.04 & 0.46 & 2.32 & 0.94 \\
    OmniDrive++~\cite{wang2024omnidrive}&CoRR'24& \checkmark &\checkmark &\checkmark & {0.14} & {0.29} & 0.55 & 0.33 & \textbf{0.01} & 0.13 & 0.78 & 0.30 \\
    
    Senna~\cite{jiang2024senna}&arXiv'24& \checkmark &- &- & 0.37 & 0.54 & 0.86 & 0.59 & 0.09 & 0.12 & 0.33 & 0.18 \\
    
    EMMA\textdagger~\cite{hwang2024emma}&arXiv'24& \checkmark &- &- &0.14 &{0.29} &0.54 &{0.32}&- &- &- &-\\
    ORION~\cite{fu2025orion} &ICCV'25& \checkmark &\checkmark & - & 0.17 & 0.31& 0.55 & 0.34& 0.05 & 0.25 & 0.80 & 0.37  \\
    OpenDriveVLA~\cite{zhou2025opendrivevla} &arXiv'25& \checkmark &\checkmark & - & 0.15 & 0.31& 0.55 & 0.33& \textbf{0.01} & 0.08 & 0.21 & \textbf{0.10}  \\
    
    \midrule
    \rowcolor[RGB]{230,242, 255}CoT4AD (\textbf{Ours})& - & \checkmark &\checkmark & - & \textbf{0.12} & \textbf{0.24}& \textbf{0.53} & \textbf{0.29}& 0.02 & \textbf{0.06} & 0.22 & \textbf{0.10}  \\
    \bottomrule
    \end{tabular}
}
\label{tab:nusc result}
\end{table*}

\subsection{Chain-of-Thought Trajectory Planning}\label{LLM}
At this stage, CoT4AD performs CoT-based planning of future driving actions, represented by a sequence of waypoints $\mathbf{V}_{ego}=\{w_1,w_2,...,w_T\}$. We adopt VLM-conditioned diffusion planning in Sec.~\ref{World Dreamer}. We follow vanilla diffusion models without latent space as illustrated in Figure~\ref{fig:arch} (b). We initialize the sampling noise from action anchors $a=\{a_k\}_{k=1}^{N}$ which is clustered by K-Means on the dataset. The diffusion transformer $f_a$ takes noisy actions $z^a_t$and conditional embedding $c_a=\mathrm{LLM}\left(\mathbf{V}_{env},\mathbf{V}_{fut},\mathbf{V}_{s},\mathbf{V}_{ego}\right)$ to predict denoised trajectories $\hat z^a$ and classification scores $\hat s^a$ as:
\begin{equation}
\{\hat s^a,\hat z^a\}=f_a(z_t,c_a,t),
\end{equation}
where $c_a$ represents the conditional embedding and $t$ represents the timestamp. The whole model, including the visual encoder, visual diffusion transformer, planning diffusion transformer and the LLM is jointly optimized  during training. To accelerate the forward process during inference, instead of explicitly generating intermediate steps and images, it directly produces $c_a$ from environment tokens and prompts as conditional embeddings. At each denoising step, the diffusion transformer takes the predicted trajectory from the previous step as input and update predictions with DDIM for the next timestep.

\section{Experiments}  \label{sec:experiment}
In this section, we first introduce two widely used datasets: nuScenes dataset \cite{caesar2020nuscenes} and Bench2Drive dataset \cite{jia2024bench2drive}. Then we evaluate our method on both datasets and compare with state-of-the-art VLA models. Furthermore, the effectiveness of our methods is explained by visual analysis. Finally, ablation studies are conducted on the Bench2Drive dataset \cite{jia2024bench2drive} to verify the effectiveness of each module.

\subsection{Datasets}

\noindent\textbf{NuScenes.} NuScenes is a widely used multimodal benchmark for perception and open-loop planning in autonomous driving, containing 1,000 20-second scenes collected in Boston and Singapore, annotated at 2 Hz (700/150/150 for train/val/test). Building on this, nuScenes-QA~\cite{qian2024nuscenes} extends it into a VQA benchmark by generating scene graphs and diverse question types (e.g., existence, counting, attribute, comparison), challenging models in multimodal reasoning.
\vspace{0.1mm}

\noindent\textbf{Bench2Drive.} Bench2Drive is a benchmark used for closed-loop E2E-AD, proposed by the Thinklab-SJTU team. It uses the CARLA V2 \cite{dosovitskiy2017carla} simulation environment to comprehensively evaluate autonomous driving systems under interactive and dynamic scenarios. For a fair baseline comparison, Bench2Drive defines a base set of 1000 clips, typically split into 950 for training and 50 for open-loop validation. Each clip corresponds to a continuous driving segment (roughly 150 meters) in a specific traffic scene. Chat-B2D is a VQA extension built on top of Bench2Drive, introduced in ORION~\cite{fu2025orion}. It provides automatically generated question-answer pairs for each driving scene in the Bench2Drive dataset, enabling semantic reasoning evaluation within closed-loop simulation.

\subsection{Evaluation Metrics}
We conduct open-loop evaluations on the nuScenes dataset using L2 distance error and average collision rate as evaluation metrics.
We conduct both open-loop and closed-loop evaluations on Bench2Drive dataset with metrics including Driving Score (DS), Success Rate (SR), Efficiency, Comfortness, and Multi-Ability. 

\subsection{Implementation Details}
In experiments, we implemented our method using the PyTorch deep learning framework on 8 NVIDIA RTX A800 GPUs with 80GB memory. The network was trained with a stochastic gradient descent (SGD)~\cite{gower2019sgd} optimizer, initialized with a learning rate of $1e^{-4}$, momentum of 0.9, and weight decay of $1e^{-4}$. For the backbone feature extraction, we adopted EVA-CLIP~\cite{fang2024eva} with pretrained model. LLaMA-3~ is employed in CoT4AD. The training diffusion schedule is truncated by 50/1000 to diffuse the latent images, while during inference, we use only 2 denoising steps for prediction.  During training, data augmentations are applied to input images, which are first resized to a resolution of 640 × 640.

\begin{table*}
\centering
\caption{Results of E2E-AD Methods on Bench2Drive under base set. C/L refers to camera/LiDAR. Avg. L2 is averaged over the predictions in 2 seconds under 2Hz, similar to UniAD. * denote expert feature distillation. NC: navigation command, TP: target point, DS: Driving Score, SR: Success Rate.
\label{tab: b2d result1}}

\resizebox{0.85\textwidth}{!}
{
\begin{tabular}{l c c c  >{\columncolor{gray!10}}c >{\columncolor{gray!10}}c c c   }
\toprule
Method & {Reference} &{*}{Condition} &{Modality}  &  DS$\uparrow$ & SR(\%)$\uparrow$ &Efficiency$\uparrow$ & Comfortness$\uparrow$ \\ 
\hline

TCP*~\cite{wu2022trajectory}& NeurIPS'22 & TP & C   &  40.70     & 15.00  & 54.26 & 47.80     \\
TCP-ctrl*~\cite{wu2022trajectory} &NeurIPS'22  & TP    & C   &  30.47    & 7.27  & 55.97 & \textbf{51.51}   \\
TCP-traj*~\cite{wu2022trajectory} & NeurIPS'22& TP & C   &  59.90     & 30.00 & 76.54 & 18.08      \\
TCP-traj w/o distillation~\cite{wu2022trajectory} & NeurIPS'22  & TP   & C    &  49.30     & 20.45  & 78.78 & 22.96     \\
ThinkTwice*~\cite{jia2023think}  & CVPR'23 & TP & C   & 62.44     & 31.23  & 69.33 & 16.22    \\
DriveAdapter*~\cite{jia2023driveadapter} &ICCV'23 & TP & C\&L   &  64.22   & 33.08 & 70.22 & 16.01     \\  

\midrule

AD-MLP~\cite{zhai2023rethinking} & arXiv'23 &  NC &C    & 18.05     &  0.00  & 48.45 &   22.63  \\
UniAD-Tiny~\cite{hu2023planning} & CVPR'23 & NC & C   & 40.73    & 13.18 & 123.92 & 47.04    \\
UniAD-Base~\cite{hu2023planning} & CVPR'23 & NC  & C  & 45.81     & 16.36 & 129.21 & 43.58    \\

VAD~\cite{jiang2023vad}& ICCV'23 & NC & C  &42.35     & 15.00 & 157.94 & 46.01  \\ 

GenAD~\cite{zheng2024genad} & ECCV'24 & NC &C & 44.81 & 15.90 &  -& - \\

MomAD~\cite{song2025don} & CVPR'25 & NC & C & 44.54	&16.71&	\textbf{170.21}&	48.63\\
DriveTransformer-Large~\cite{jia2025drivetransformer} & ICLR'25 & NC   & C & 63.46     & 35.01 & 100.64 & 20.78   \\ 
ORION~\cite{fu2025orion}  & ICCV'25 & NC & C & 77.74  & {54.62} & 151.48& 17.38    \\

\midrule
\rowcolor[RGB]{230, 242, 255}Cot4AD (\textbf{Ours})  & - & NC & C & 80.24  & 55.22 & 153.33& 46.78    \\
\rowcolor[RGB]{230, 242, 255}Cot4AD-CoT (\textbf{Ours})  & - & NC & C & \textbf{81.22} & \textbf{55.78} & 149.28& 47.40    \\

\bottomrule
\end{tabular}
}
\end{table*}

\begin{figure*}[!htb]
	\centering
	\includegraphics[width=\textwidth]{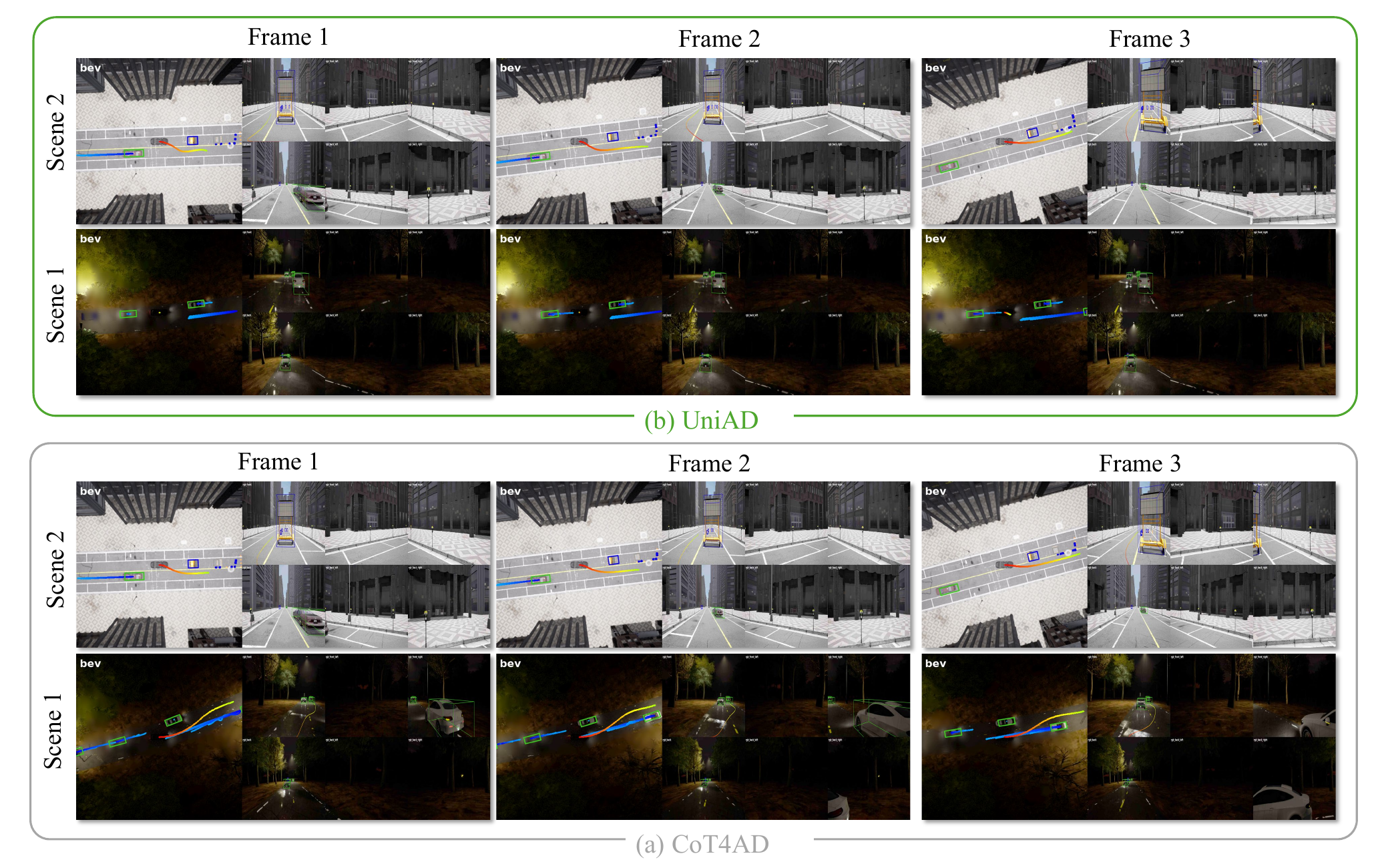}
	\caption{Qualitative results of CoT4AD on the Bench2Drive closed-loop evaluation set.}
	\label{fig:vis_b2d}
    \vspace{-0.4cm}
\end{figure*}

\subsection{Comparison with the State-of-Art Methods}

\noindent \textbf{NuScenes.} Table~\ref{tab:nusc result} reports the performance of state-of-the-art E2E-AD methods on the NuScenes dataset under UniAD metrics. CoT4AD achieves competitive results with L2 errors of 0.12 m, 0.24 m, and 0.53 m for 1s, 2s, and 3s horizons, respectively (avg. 0.29 m), clearly outperforming recent VLM-based counterparts such as OpenDriveVLA and EMMA. In terms of safety, CoT4AD maintains a notably low collision rate of 0.10\%, demonstrating robust and reliable trajectory prediction.
Unlike methods that incorporate planner trajectories, CoT4AD relies on ego BEV status and camera images as input, yet consistently achieves lower prediction errors and collision rates than OmniDrive++. This highlights the strong reasoning ability of the proposed chain-of-thought reasoning, which enhances the model’s capability to predict driving trajectories in complex scenes.

\noindent \textbf{Bench2Drive.} Table \ref{tab: b2d result1} presents the closed-loop results of various E2E-AD methods on the Bench2Drive benchmark. CoT4AD indicates that the model outputs the trajectories end-to-end, while CoT4AD-CoT denotes that the trajectories are produced through step-by-step CoT reasoning. Our proposed CoT4AD and CoT4AD-CoT models achieve the best overall performance. 
Specifically, CoT4AD-CoT achieves the highest Driving Score with 81.22 and Success Rate with 55.78\%, surpassing strong baselines such as ORION and DriveTransformer-Large.
In terms of Efficiency and Comfortness, our method achieves a balanced performance—demonstrating smoother and more human-like control behavior compared with transformer-based approaches that tend to produce abrupt maneuvers.
Notably, both versions of CoT4AD rely solely on camera inputs and navigation commands, yet outperform models that incorporate additional LiDAR or expert distillation signals. These results highlight the effectiveness of our chain-of-thought reasoning framework in enhancing decision consistency and improving driving reliability under diverse scenarios.
For more detailed experiment results, please see in the Appendix.

\subsection{Qualitative Results}

To further demonstrate the planning capability of our approach in complex scenarios, we conduct a qualitative comparison between CoT4AD and UniAD. As shown in Figure~\ref{fig:vis_b2d}, two representative driving scenes are selected, each containing three consecutive frames to illustrate the models’ decision-making processes. In Scene 1, the model is instructed to change lanes to avoid an obstacle ahead. Although both methods produce reasonable lane-change trajectories, CoT4AD generates a smoother and more continuous trajectory, maintaining a larger safety margin from the obstacle in frame 3.  In Scene 2, the model is instructed to perform an overtaking maneuver. UniAD fails to anticipate the overtaking intent early enough, resulting in insufficient acceleration and limited overtaking distance, which prevents it from completing the maneuver safely. In contrast, CoT4AD is capable of recognizing the intent earlier and planning proactively, maintaining a closer following distance while accelerating and changing lanes smoothly to complete the overtaking successfully. This demonstrates that CoT4AD possesses stronger temporal reasoning and high-level semantic understanding, enabling more human-like and context-aware driving behaviors. Overall, these qualitative results highlight the advantages of the proposed Chain-of-Thought reasoning: it not only improves trajectory smoothness and safety but also enhances robustness and decision interpretability in complex traffic interactions. For additional qualitative visualizations and videos, please refer to the Appendix.

\subsection{Ablation Studies}

\begin{table}[!t]

\setlength{\tabcolsep}{1.8mm}
\caption{Ablation results of tokenizer on Bench2Drive. DS and SR denote Driving Score and Success Rate.
}
\centering
\resizebox{0.35\textwidth}{!}
{
\begin{tabular}{c ccc |cc}
\toprule
\multirow{2}{*}{\textbf{ID}} &\multirow{2}{*}{\textbf{$T_{map}$}} &\multirow{2}{*}{\textbf{$T_{obj}$}}  &\multirow{2}{*}{\textbf{$T_{bev}$}}    &
\multicolumn{2}{c}{\textbf{Closed-loop}}\\
& & & & \textbf{DS $\uparrow$} & \textbf{SR (\%)  $\uparrow$}   \\
\midrule
1 &\yes&\no& \no& 44.21  & 20.12 \\
2 &\no&\yes& \no& 56.28  & 38.02   \\
3 &\yes&\yes& \no& 79.77  & 50.88  \\
4 &\no&\no& \yes& 69.92  & 39.64    \\
5 &\yes&\yes& \yes & \textbf{80.24}  & \textbf{55.22}   \\
\bottomrule
\end{tabular}}
\label{tab:ablation studies 1}

\vspace{-2mm}
\end{table}

\textbf{Effectiveness of Perception Tokenizers.} Table \ref{tab:ablation studies 1} shows the detailed ablations of each tokenizer. We observe that using Tokenizer based on perception labels (ID 1 and ID 2) achieves relatively strong performance. For example, using only $T_{map}$ or $T_{obj}$ already significantly improves the success rate (SR), indicating that perception labels provide valuable guidance for planning. Tokenizer based directly on visual features (ID 3 \vs ID 4) also achieves good performance but is less effective than the perception-based tokenizer (50.88 \vs 39.64). Combining both perception and visual features further enhances performance, achieving the best closed-loop metrics. This indicates that although limited perception labels can provide certain environmental information, image features retain more complete semantic information, and combining both can enhance the model's overall performance.

\begin{figure}[!htb]
	\centering
	\includegraphics[width=0.9\linewidth]{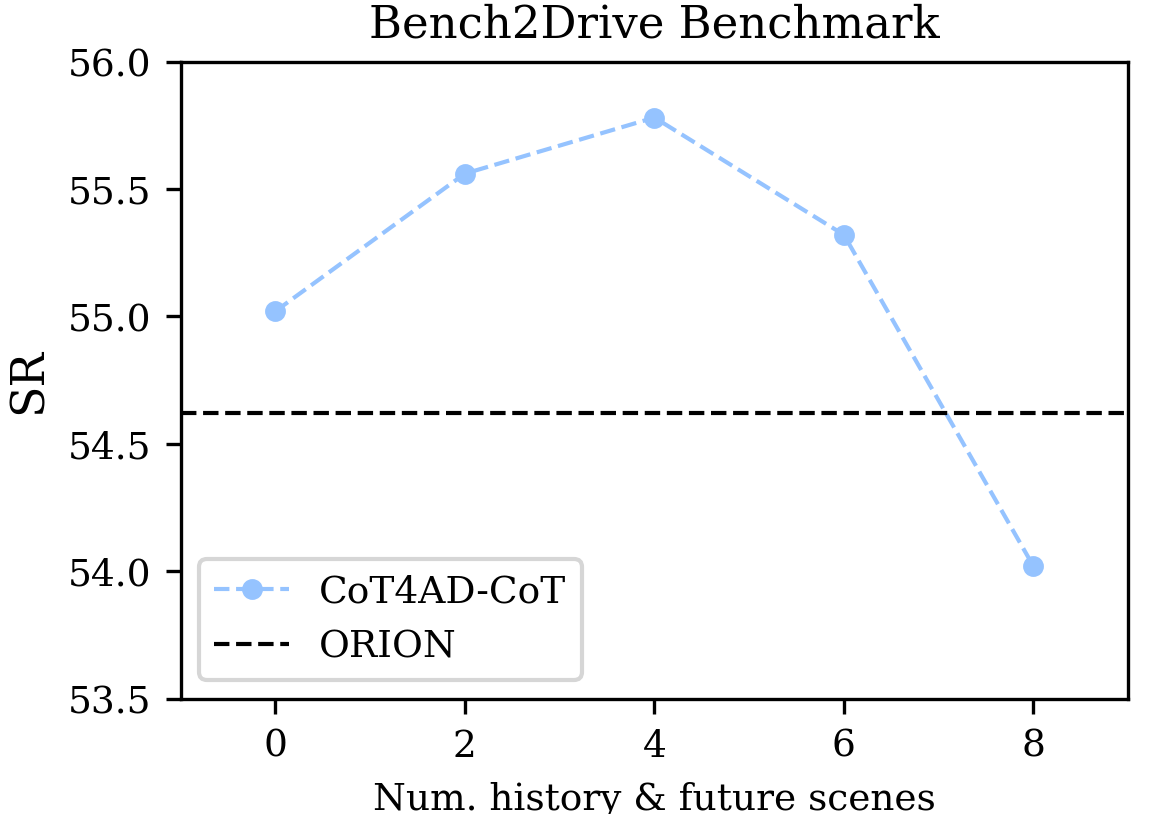}
	\caption{Ablation on the number of predicted future scenes}
    \vspace{-3mm}
	\label{fig:diffusion ablation}
\end{figure}

\noindent\textbf{Effectiveness of CoT Designs.} Table~\ref{tab:ablation studies 2} shows the detailed ablations of each step in the introduced CoT pretraing. Removing basic perception (ID 1 \vs ID 2) leads to a significant drop in both DS (66.44 \vs 59.89) and SR (36.86 \vs 27.95), indicating that visual perception features are essential for safe and successful driving. Similarly, removing the VQA module (ID 1 \vs ID 3) reduces a smaller DS (64.38 \vs 59.89) and SR (33.22 \vs 27.95), showing that reasoning about visual questions provides useful contextual information for driving decision-making. However, the future diffusion has a greatest impact on performance. With only the future module enabled (ID 4), both DS and SR increase substantially (DS: 72.38, SR: 45.22), highlighting the importance of anticipating future states in closed-loop control. This phenomenon may be due to the fact that, compared with limited annotated labels, the model learns richer perception representations from future predictions through self-supervised learning. As a result, combining all components (ID 5) achieves the best performance (DS: 80.24, SR: 55.22), demonstrating that perception, VQA reasoning, and future prediction complement each other to enhance driving performance.

\begin{table}[!t]

\setlength{\tabcolsep}{1.8mm}
\caption{Ablation results of CoT pre-training. DS and SR denote Driving Score and Success Rate.
}
\centering
\resizebox{0.4\textwidth}{!}
{
\begin{tabular}{c ccc |cc}
\toprule
\multirow{2}{*}{\textbf{ID}} &\multirow{2}{*}{\textbf{Perception}} &\multirow{2}{*}{\textbf{VQA}}  &\multirow{2}{*}{\textbf{Future}}    &
\multicolumn{2}{c}{\textbf{Closed-loop}}\\
& & & & \textbf{DS $\uparrow$} & \textbf{SR (\%)  $\uparrow$}   \\
\midrule
1 &\no&\no& \no& 59.89  & 27.95 \\
2 &\yes&\no& \no& 66.44  & 36.86   \\
3 &\no&\yes& \no& 64.38  & 33.22  \\
4 &\no&\no& \yes& 72.38  & 45.22    \\
5 &\yes&\yes& \yes & \textbf{80.24}  & \textbf{55.22}   \\
\bottomrule
\end{tabular}}
\label{tab:ablation studies 2}

\vspace{-2mm}
\end{table}

\noindent\textbf{Effectiveness of Future Scenes Prediction}. Figure \ref{fig:diffusion ablation} presents a detailed ablation study on the impact of the number of predicted future scenes used in the model. As the number of future scenes increases from 0 to 8, the model's performance improves, reaching a peak with 4 predicted future scenes, achieving a SR of 55.78\%. However, when the number of predicted future scenes exceeds this optimal threshold, the model's performance begins to degrade. This decline is attributed to the excessive inclusion of future information, which results in overloading the model and introducing confusion, thereby diminishing the SR. These results highlight the importance of balancing the incorporation of future scene predictions.


\section{Conclusion and Future Work} \label{sec:conclusion}
In this paper, we propose CoT4AD, a model that performs explicit chain-of-thought reasoning in autonomous driving scenarios. By adopting a multi-step reasoning process consisting of perception–VQA–diffusion–planning, the model achieves better alignment across the visual, reasoning, and action spaces, enabling smoother and more accurate planning in driving tasks. Experiments conducted on two datasets demonstrate that CoT4AD outperforms existing state-of-the-art methods, validating the effectiveness of our proposed approach.  Although CoT4AD performs well under explicit reasoning, it is constrained by the computational complexity of multi-step CoT reasoning and the training instability of diffusion models. In future work, we plan to explore more efficient CoT reasoning mechanisms to enable real-time autonomous driving.

\newpage
{
    \small
    \bibliographystyle{ieeenat_fullname}
    \bibliography{main}
}


\end{document}